\documentclass[10pt, a4paper]{article}

\usepackage{lrec-coling2024} % this is the new style
\usepackage{subfigure}
\usepackage{latexsym}
\usepackage{graphicx}
\usepackage{multicol}
\usepackage{multirow}
\title{Uncovering the Potential of ChatGPT for Discourse Analysis in Dialogue: An Empirical Study}
\name{Yaxin Fan\textsuperscript{1}\sthanks{ \quad Work done during a visiting student at CUHKSZ NLP group.}, Feng Jiang\textsuperscript{2,3}\sthanks{ \quad Corresponding author.}, Peifeng Li\textsuperscript{1}, Haizhou Li\textsuperscript{2}} 

\address{\textsuperscript{1}School of Computer Science and Technology, Soochow University, Suzhou, China \\
        \textsuperscript{2}Shenzhen Research Institute of Big Data, School of Data Science, \\ The Chinese University of Hong Kong, Shenzhen (CUHK-Shenzhen), Guangdong, China \\
        \textsuperscript{3}School of Information Science and Technology,\\
        University of Science and Technology of China, Hefei, China \\
         yxfansuda@stu.suda.edu.cn, pfli@suda.edu.cn\\
         \{jeffreyjiang,haizhouli\}@cuhk.edu.cn\\}

\abstract{Large language models, like ChatGPT, have shown remarkable capability in many downstream tasks, yet their ability to understand discourse structures of dialogues remains less explored, where it requires higher level capabilities of understanding and reasoning. In this paper, we aim to systematically inspect ChatGPT's performance in two discourse analysis tasks: topic segmentation and discourse parsing, focusing on its deep semantic understanding of linear and hierarchical discourse structures underlying dialogue. To instruct ChatGPT to complete these tasks, we initially craft a prompt template consisting of the task description, output format, and structured input. Then, we conduct experiments on four popular topic segmentation datasets and two discourse parsing datasets. The experimental results showcase that ChatGPT demonstrates proficiency in identifying topic structures in general-domain conversations yet struggles considerably in specific-domain conversations. We also found that ChatGPT hardly understands rhetorical structures that are more complex than topic structures. Our deeper investigation indicates that ChatGPT can give more reasonable topic structures than human annotations but only linearly parses the hierarchical rhetorical structures. In addition, we delve into the impact of in-context learning (e.g., chain-of-thought) on ChatGPT and conduct the ablation study on various prompt components, which can provide a research foundation for future work. The code is available at \url{https://github.com/yxfanSuda/GPTforDDA}.
 \\ \newline \Keywords{Large Language Model, Dialogue Topic Segmentation, Dialogue Discourse Parsing, Chain-of-Thought} }

\begin{document}

\maketitleabstract

\section{Introduction}

With the development of generative models, large language models (e.g., ChatGPT) have exhibited remarkable capability on various natural language generation (NLG)\cite{jiao2023chatgpt, zhang2023extractive, zhang2023benchmarking, yang2023exploring} and understanding (NLU) \citep{wei2023zeroshot, yuan2023zeroshot, hu2023zeroshot, gao2023exploring} tasks. Despite this progress, there remains an absence of a qualitative and quantitative evaluation of ChatGPT on dialogue discourse analysis. Such an evaluation is vital to uncover the potential of ChatGPT for deep semantic understanding of conversations.

\begin{figure}[ht!]
	\centering  
        \includegraphics[width=\linewidth]{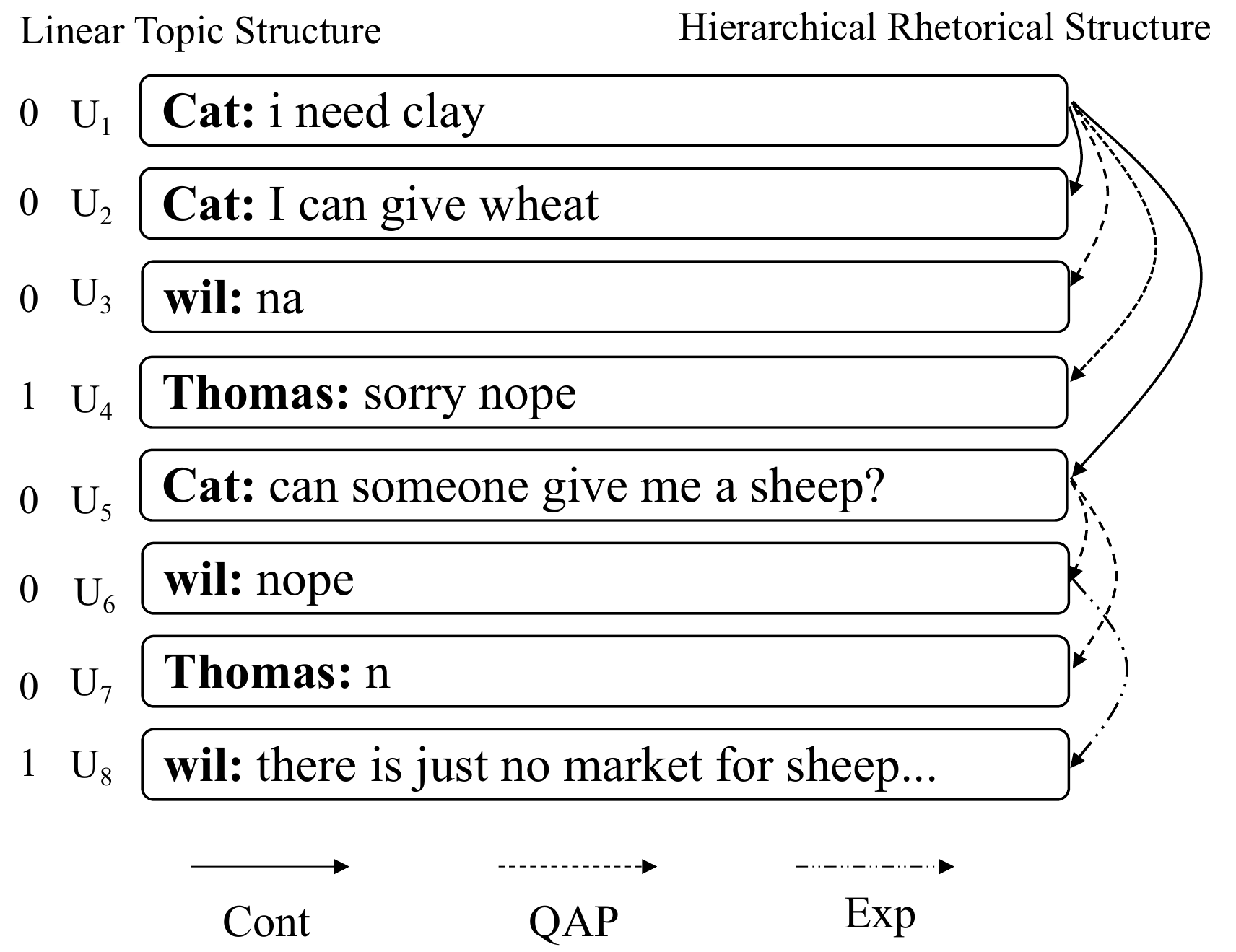}
        \caption{A dialogue from the STAC \citep{asher2016discourse} dataset, consisting of seven utterances $U_1$-$U_7$ and three speakers \emph{Cat}, \emph{wil}, and \emph{Thomas}. Dialogue topic segmentation aims to reveal the linear topic structure by dividing the dialogue into several topical pieces and '1' indicates the end of a topic. Dialogue discourse parsing aims to reflect hierarchical rhetorical structure by establishing discourse links of utterance pairs according to discourse relations, where \emph{Cont}, \emph{QAP}, and \emph{Exp} is short for \emph{Continuation}, \emph{Question-answer\_pair}, and \emph{Explanation}, respectively. }
         \label{description_task}
          \vspace{-0.3cm}
    \end{figure}

Dialogue discourse analysis plays a crucial role in natural language processing (NLP) by revealing the underlying topic, coherence, and rhetorical structures in a dialogue. Most previous work in this field mainly focuses on dialogue topic segmentation \cite{lin2023multigranularity, gao2023unsupervised, lin2023topic, xing-carenini-2021-improving,xie-etal-2021-tiage-benchmark} and discourse parsing \cite{chi-rudnicky-2022-structured,fan2022distance,yu2022speaker,ijcai2021-543,shi2019deep}, aiming to study the linear topic structures and hierarchical rhetorical structures, respectively, as depicted in Figure~\ref{description_task}. Considering ChatGPT as the most recent language generation capability, the potential of its discourse structure understanding proficiency remains largely uncharted. Unlike some NLP tasks that only require a shallow semantic understanding to extract the output from the input, dialogue discourse analysis poses a unique challenge for LLMs, which requires a deeper semantic understanding to derive the latent discourse structures.
Therefore, our study delves into the performance of ChatGPT on topic segmentation and discourse parsing to explore the capability of ChatGPT in deep semantic understanding, including linear topic structure and hierarchical rhetorical structure.

To this end, we first crafted the prompt consisting of three components: task description, output format, and structured input. The task description tells ChatGPT what needs to be accomplished, the output format instructs ChatGPT to output in a specified format that can easily extract linear/hierarchical structure for evaluation, and the structured input provides ChatGPT with organized content that needs to be analyzed. Then, we conducted two times experiments on four popular topic segmentation datasets and two discourse parsing datasets and reported the average performance. 

Experimental results showcased that ChatGPT has a good understanding of linear topic structures in the general domain but struggles to understand the topic structure of the specific domain. Besides, ChatGPT can hardly understand the hierarchical rhetorical structures. In-depth analysis reveals that ChatGPT can give more reasonable topic structures than human annotations but only parses the hierarchical rhetorical structures linearly.

In addition, attempting to enhance the abilities of ChatGPT for deep semantic understanding, we explored the effect of In Context Learning (ICL).  The results showcased that ICL could not facilitate ChatGPT to understand linear structures but improve the abilities of ChatGPT to understand hierarchical structures.  Notably, the chain-of-thought method of ICL contributes the most.

Furthermore, we conducted ablation experiments to explore the role of various prompt components. The ablation results reveal that the output format plays the most important role. This provides some insights into crafting prompts for studying the dialogue discourse analysis.

Finally, we studied the instruction-following abilities of ChatGPT and found that ChatGPT can not fully follow the instructions on all datasets. This indicates that the robustness of ChatGPT is still an issue of concern.

We hope that our study can provide a solid foundation for the research of dialogue discourse analysis in the future.

\section{Related work}
\subsection{Evaluation of ChatGPT}
Recently, some works have evaluated the performance of ChatGPT on a series of downstream tasks, including machine translation \citep{jiao2023chatgpt}, summarization \citep{zhang2023extractive, yang2023exploring}, information extraction \citep{wei2023zeroshot, han2023information,yuan2023zeroshot, hu2023zeroshot,gao2023exploring}, and other NLP tasks \citep{pu2023chatgpt, susnjak2023applying, qin2023chatgpt}. Most of the previous studies explore the capabilities of ChatGPT for shallow semantic understanding that obtaining the output according to the mapping relation between input and output or directly extracting the content from the input. Different from these tasks, dialogue discourse analysis requires ChatGPT to have deeper semantic understanding ability to deduce the discourse structures underlying dialogue.

% Detecting the performance of ChatGPT in a specific task can be mainly divided into three aspects: exploring the impact of different inputs, exploring the impact of task execution methods, and XXX.
%The key to exploring the abilities of ChatGPT on a particular task is to design an appropriate prompt. We designed the prompt consisting of three components: task description, output format, and structured input to investigate the potential of ChatGPT in dialogue discourse analysis. 
\subsection{Dialogue Topic Segmentation}
Previous work on dialogue topic segmentation mainly identifies topic boundaries by studying the local coherence of consecutive utterances, which is divided into two types: unsupervised and supervised.
The unsupervised methods \citep{song2016dialogue, Xu_Zhao_Zhang_2021,xing-carenini-2021-improving} mainly first train a coherence model to assess the similarity of consecutive utterances, then a global segmentation algorithm, such as TextTiling \citep{hearst1997text}, is adopted to identify the topic boundaries.
And those supervised approaches \cite{xie-etal-2021-tiage-benchmark, lin2023topic, lin2023multigranularity} mainly adopt sequence labeling to determine the topic boundaries.
Different from the previous work with a 0/1 sequences as the output format, we instruct ChatGPT to output the consecutive utterances within the same topic in a dialogue. 

\subsection{Dialogue Discourse Parsing}
Traditional work on dialogue discourse parsing mainly studies the coherence between any two utterances and then adopts a global decoding (e.g., maximum spanning trees) to parse the hierarchical rhetorical structure.
The work can be divided into two types, i.e., model augmentation and data augmentation. Those model augmentation \cite{shi2019deep, ijcai2021-543, fan2022distance, chi-rudnicky-2022-structured} approaches mainly design various sophisticated encoding or decoding methods for discourse parsing.  
Data augmentation approaches \cite{yang2021joint, he2021multi, liu2021improving, yu2022speaker, fan-etal-2023-improving} mainly integrate the data of related tasks to facilitate dialogue discourse parsing. 
Different from the previous work that adopts the adjacency matrix as the output format, we instruct ChatGPT to output the rhetorical structures in the format of a sparse matrix.

\begin{table*} \scriptsize
\centering
\setlength{\tabcolsep}{1mm}{
\begin{tabular}{ccc}
\hline
\textbf{Elements}         & \textbf{Dialogue Topic Segmentation} & \textbf{Dialogue Discourse Parsing} \\                                                                      \hline                                                                                  
                                                                             
\textbf{Task Description}        & \begin{tabular}[l]{@{}l@{}}Please identify several topic boundaries for the following \\ dialogue and each topic consists of several consecutive \\utterances. 

\end{tabular}   &\begin{tabular}[l]{@{}l@{}}
According to the Segmented Discourse Rhetorical Theory, \\the rhetorical structure of a dialogue can be represented by \\ a directed acyclic graph, where nodes are utterances \\and edges are the following 16 relations:\\
$[$"Comment", "Clarification-Question","Elaboration",\\ "Acknowledgement", "Explanation", "Conditional", \\   
"Question-Answer pairs", "Alternation","Question-Elabo\\ration",
"Result", "Background", "Narration",\\
"Correction", "Parallel", "Contrast",\\ "Continuation"$]$\\
\end{tabular}

\\ \hline

\textbf{Output Format}        & \begin{tabular}[l]{@{}l@{}}please output in the form of \{'topic i':{[}{]}, ... ,'topic j':{[}{]}\}, \\where the elements in the list are the index of the \\ consecutive utterances within the topic, and output \\ even if there is only one topic.
\end{tabular}    & \begin{tabular}[l]{@{}l@{}}please annotate the rhetorical structure of the following \\ dialogue and represent it in the form of [index1, index2, \\'relation'], where index1 and index2 are the index of \\ two utterances, and the 'relation' is one of the \\ above relations to connect the two utterances.
\end{tabular} 
\\ \hline

\textbf{Structured Input}        &  \multicolumn{2}{c}{\begin{tabular}[l]{@{}l@{}}
$0: U_1$\\
$1: U_2$\\
$\cdots$\\
$n: U_n$\\
         
\end{tabular}} \\ \hline

\end{tabular}}
\caption{Prompt template for Dialogue Discourse Analysis.}
\label{promptsDDA}
\end{table*}

\section{ChatGPT for Dialogue Discourse Analysis}
\subsection{Prompt Template Design}
The key to evaluating the performance of ChatGPT on specific tasks is to design appropriate prompts \citep{jiao2023chatgpt,wei2023zeroshot,han2023information}. 
Unlike other tasks, e.g., translation, information extraction, etc., which require a simple task description, guiding ChatGPT in dialogue discourse analysis requires not only the task description but also the output format of discourse structure for evaluation.
To this end, we crafted the prompt template consisting of three components: task description, output format, and structured input, as shown in Table~\ref{promptsDDA}.

\paragraph{Task Description}
The task description guides ChatGPT to understand and complete the task as required. For each task, we describe the goal of the task, such as identifying several boundaries for dialogue topic segmentation, to instruct ChatGPT to understand and complete the task.

\paragraph{Output Format}
The output format instructs ChatGPT to output in a specified format that can easily extract linear/hierarchical structure for evaluation. 
Specifically, the linear topic structure is output in the format of a Python dictionary, where the key is the topic indication and the value is a list containing the index of consecutive utterances. The hierarchical rhetorical structure is output in the format of a sparse matrix.  The Python dictionary and sparse matrix make it simple to extract the linear and hierarchical structures for assessment and analysis.
% The linear and hierarchical structures can be easily extracted from the Python dictionary and sparse matrix for evaluation and analysis.
% We provide an example of the output format for ChatGPT and describe the role of each element in this example to help ChatGPT to generate in the specified output format.

\paragraph{Structured Input}
The structured input provides ChatGPT with organized content that needs to be analyzed. We number each utterance in the conversation and feed them into ChatGPT line by line.

\begin{figure}[t]
	\centering  %图片全局居中
	
	\subfigure[Inter-topic processing.]{
	\includegraphics[width=0.6\linewidth]{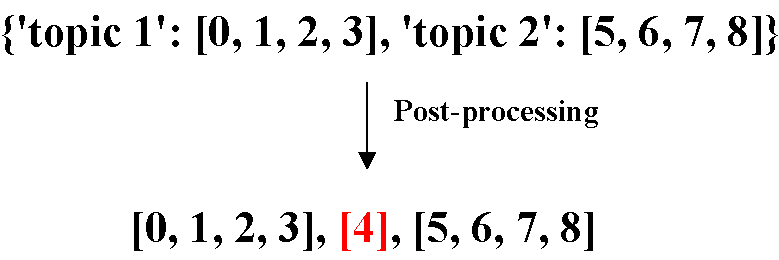}
        % \caption{sss}
        % \label{figure3_1}
        }
	\subfigure[Intra-topic processing.]{
        \includegraphics[width=0.6\linewidth]{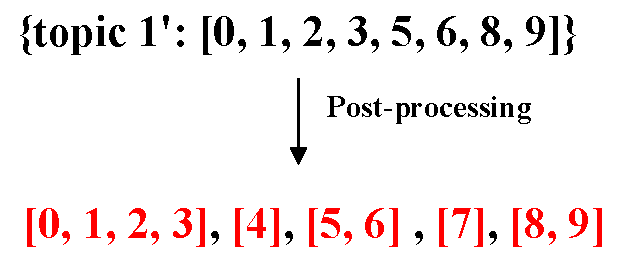}
        % \caption{sss}
        % \label{figure3_2}
        }
        \caption{Post-processing for dialogue topic segmentation.}
        \label{post_processing_figure}
    \end{figure}
\subsection{Post-processing}
Since ChatGPT is a generative model, the output can not always follow the format specified. For these outputs that do not follow the specified format, we have to conduct post-processing for evaluation.

For dialogue topic segmentation, two types of output need post-processing as shown in Figure~\ref{post_processing_figure}. First is the inter-topic processing that lacks utterances between adjacent topics and we treat the lacking utterances as an independent topic. The second is the intra-topic processing where the utterances with a topic are discontinuous and we divide the utterances within the topic into several sub-topics by the principle of maximum continuous utterance.

For dialogue discourse parsing, the relations provided by ChatGPT are occasionally not among the candidate relations, and we randomly choose one from the candidate relations.

\begin{table*}[htbp]\scriptsize
\centering
\renewcommand\arraystretch{1}
\setlength{\tabcolsep}{2.5mm}{

\begin{tabular}{ccccccc|cc}
\hline
\multirow{4}{*}{\textbf{Method}} & \multicolumn{6}{c}{\textbf{General-domain}}                                                    & \multicolumn{2}{c}{\textbf{Specific-domain}} \\
                        & \multicolumn{2}{c}{\textbf{DialSeg711}} & \multicolumn{2}{c}{\textbf{CNTD}} & \multicolumn{2}{c}{\textbf{TIAGE}} & \multicolumn{2}{c}{\textbf{ZYS}}             \\
                          & \multicolumn{2}{c}{(Daily booking service)} & \multicolumn{2}{c}{(Chitchat w/ background)} & \multicolumn{2}{c}{(Chitchat w/o background)} & \multicolumn{2}{c}{(Banking expertise)}             \\
                       & \textbf{$P_k$($\downarrow$)}            & \textbf{$F_1$($\uparrow$) }      & \textbf{$P_k$($\downarrow$)}         & \textbf{$F_1$($\uparrow$) }  & \textbf{$P_k$($\downarrow$) }           & \textbf{$F_1$($\uparrow$)}       & \textbf{$P_k$($\downarrow$)}         & \textbf{$F_1$($\uparrow$)} \\ \hline
TextTiling  & 40.44        & 60.80    & 51.36              & 46.84   & 47.27    & 45.57     & 45.86     & 48.50   \\
GreedySeg   & 50.95        & 40.10    & 53.81               & 53.36    & 52.63    & 49.47    & 44.12     & 50.20 \\
TeT+CLS     & 40.49        & 61.00     & 43.01  &50.20      & \textbf{40.49}       & \textbf{61.00}      & 43.01     & 50.20   \\
 UPCS       & 26.80        & 77.60     & 46.11   & 58.18   & 47.19    & 58.63     & \textbf{40.99}     & \textbf{52.10} \\ \hline
ChatGPT        & \textbf{10.56(0.18) }                   & \textbf{89.42(0.08)}                & \textbf{27.08(1.00)}                    &  \textbf{77.36(0.35)}                 &42.35(2.31)                        &  \textbf{61.31(1.87) }                    & 56.19(0.29)                       & 49.10(0.04) \\
Ratio@SOTA    & 253.79\%                    & 115.23\%                  &170.27\%                     & 132.97\%                  & 95.61\%    & 100.51\%                       & 72.95\%                       & 94.24\%
 \\ \hline
\end{tabular}}

\caption{Performance comparison between ChatGPT and unsupervised baselines on dialogue topic segmentation. }
\label{TS_results}
\end{table*}

\begin{table}[]\scriptsize
\centering
\begin{tabular}{ccccc}
\hline
\multirow{2}{*}{\textbf{Method}}        & \multicolumn{2}{c}{\textbf{CNTD}}                            & \multicolumn{2}{c}{\textbf{TIAGE}}                           \\
                                & \textbf{$P_k$($\downarrow$)}            & \textbf{$F_1$($\uparrow$) }      & \textbf{$P_k$($\downarrow$)}         & \textbf{$F_1$($\uparrow$)}                         \\
                               \hline
BERT                           & -                      & 80.80                       & -                    & 66.60                       \\
T5                             &  -                     & 81.10                       & -                    & 73.90                       \\
MGP                            & -                      & \textbf{84.70}                       &  -                   & \textbf{76.20}                       \\ \hline
ChatGPT                        & 27.08                 & 77.36                       & 42.35                 & 61.31                       \\
\multicolumn{1}{c}{Ratio@SOTA} & \multicolumn{1}{c}{-} & \multicolumn{1}{c}{91.33\%} & \multicolumn{1}{c}{-} & \multicolumn{1}{c}{80.46\%} \\
\hline
\end{tabular}
\caption{Performance comparison between ChatGPT and supervised baselines on dialogue topic segmentation.}
\label{TS_results_supervised}
\end{table}

\section{Experiments}
\subsection{Experimental Setup} 
For all experiments, we adopted the \emph{gpt-3.5-turbo-0301} version of ChatGPT, and all hyperparameters are set to default as recommended by OpenAI. Since ChatGPT may generate empty responses (i.e., empty strings) as the result of network error or API request overloads, we resubmit the request until ChatGPT provides non-empty responses. All experiments for each task are conducted two times, and we report the mean and standard deviation values to alleviate the randomness of ChatGPT.

For dialogue topic segmentation, we adopt $P_k$ error score \citep{beeferman1999statistical} and Macro $F_1$ score metrics. $P_k$ is to calculate the overlap rate between predicted and reference pieces and lower scores indicate better performance. $F_1$ is to measure the performance of the binary prediction. For dialogue discourse parsing, we adopt Link $F_1$ and Link\&Rel $F_1$ metrics. Link $F_1$ metric evaluates the capability of link prediction only, and the Link\&Rel $F_1$ metric evaluates the capability of link and relation are all correct.

\subsection{Datasets}
\subsubsection{Dialogue Topic Segmentation}
We mainly evaluate the performance of ChatGPT on three general-domain and one specific-domain dialogue topic segmentation datasets. 
\textbf{General-domain dataset}: \textbf{DialSeg711} \citep{Xu_Zhao_Zhang_2021}: it is a synthetic dataset about reservations that consists of 711 English dialogues for unsupervised evaluation. Topics of the dataset are mainly about booking tickets, hotels, taxis, etc. \textbf{CNTD} \citep{lin2023topic}: it is a real-world Chinese chitchat dataset that consists of 1041, 134, and 133 conversations for training, validating, and testing, respectively. Participants always engage in a conversation around a given news report. \textbf{TIAGE} \citep{xie-etal-2021-tiage-benchmark}: it is a real-world English chitchat dataset that consists of 300, 100, and 100 dialogues for training, validating, and testing, respectively. Unlike CNTD, participants of the dataset engage in conversations aimlessly. \textbf{Specific-domain dataset}: \textbf{ZYS} \citep{Xu_Zhao_Zhang_2021}: it is a real-world Chinese dataset about banking consultation that consists of 505 conversations for unsupervised evaluation. The conversations in this dataset are always about banking expertise.

\subsubsection{Dialogue Discourse Parsing}
We evaluate the performance of ChatGPT on two datasets \textbf{STAC} \citep{asher2016discourse} and \textbf{Molweni} \citep{li-etal-2020-molweni}. STAC is collected from an online game \emph{The Settlers of Catan}, which contains 1,062 and 111 dialogues for training and testing, respectively. Molweni is based on Ubuntu Chat \citep{lowe2015ubuntu}, which contains 9,000, 500, and 500 instances for training, validating, and testing, respectively. Both datasets define 16 relation types. We follow the previous work and evaluate the performance of ChatGPT on the testing set of all datasets.

\subsection{Baseline}
% \paragraph{Dialogue Topic Segmentation}
The baselines for dialogue topic segmentation are two types: unsupervised and supervised. Unsupervised baselines: 1) \textbf{TextTiling} \citep{hearst1997text}: it is a traditional and common method that uses word frequencies to measure the similarity among the utterances. 2) \textbf{GreedySeg} \citep{Xu_Zhao_Zhang_2021}: This method greedily determines segment boundaries based on the similarity of adjacent utterances computed from the output of the pre-trained BERT sentence encoder. 3) \textbf{TeT+CLS} \citep{Xu_Zhao_Zhang_2021}: TextTiling enhanced by the pre-trained BERT sentence encoder, by using output embeddings of BERT encoder to compute semantic similarity for consecutive utterance pairs. 4) \textbf{UPCS} \citep{xing-carenini-2021-improving}: it is a distant supervised method and trains an utterance-pair scoring model by sampling utterance pairs from distant corpora DailyDialog\citep{li-etal-2017-dailydialog} and Naturalconv \citep{Wang_Li_Zhao_Yu_2021}.
Supervised baselines: 1) \textbf{BERT} \citep{lin2023multigranularity}: it uses BERT \citep{devlin2019bert} to encode utterance pairs and train a binary classifier. 2) \textbf{T5} \citep{xie-etal-2021-tiage-benchmark}: it uses T5 \citep{raffel2020exploring} to encode utterance pairs and train a binary classifier. 3) \textbf{MGP} \citep{lin2023topic}: it proposes a prompt-based method to fully extract topic information at several granularities from dialogues.

The baselines for dialogue discourse parsing are as follows:
1) \textbf{Rule-based}: it establishes the discourse links between adjacent utterances and treats the relation with the most common type. It can be regarded as the linear representation of discourse structure.
2) \textbf{DSM} \citep{shi2019deep}: it alternately predicted the link and relation by incorporating historical structure; 3) \textbf{SSAM} \citep{ijcai2021-543}: it adopted a structure transformer and two auxiliary training signals for parsing; 4) \textbf{DAMT} \citep{fan2022distance}: it combined different decoding methods for parsing; 
5) \textbf{SSP} \citep{yu2022speaker}: it proposed a second-stage pre-trained task to enhance the speaker interaction;
6) \textbf{SDDP} \citep{chi-rudnicky-2022-structured}: it proposed to jointly optimize discourse links and relations in the dialogue and use the modified Chiu-Liu-Edmonds algorithm to generate discourse structure.

\begin{table}[]\scriptsize
\centering

\setlength{\tabcolsep}{0.5mm}{
\begin{tabular}{ccccc}
\hline
\multirow{2}{*}{\textbf{Methods}} & \multicolumn{2}{c}{\textbf{STAC}} & \multicolumn{2}{c}{\textbf{Molweni}}  \\
                 & \textbf{Link F1}   & \textbf{Link\&Rel F1}        & \textbf{Link F1}   & \textbf{Link\&Rel F1}   \\
\hline
Rule-based &60.57   &20.11              &67.56                  &25.60 \\ 
DSM   & 71.99      &  53.62             & 76.94                 &53.49     \\
SSAM  & 73.48      &  57.31             & 81.63                 &58.54 \\
SSP   & 73.00      &  57.40             & 83.70                 &59.40  \\
DAMT  & 73.64      &  57.42             & 82.50                 & 58.91 \\
SDDP  & \textbf{74.40}      & \textbf{59.60}              & \textbf{83.50}                 & \textbf{59.90}\\
\hline 
ChatGPT   &59.91(0.13)      & 25.25(0.88)       &	63.75(0.04)     & 23.85(0.06)     \\
Ratio@SOTA   &80.52\%      & 42.37\%       &	76.35\%     & 39.82\% \\
\hline
\end{tabular}
}
\caption{Performance comparison between ChatGPT and baselines on dialogue discourse parsing.}
\label{DDP_results}
\end{table}

\section{Experimental Results}

\subsection{Results on Dialogue Topic Segmentation} 

Table~\ref{TS_results} shows the performance comparison between ChatGPT and unsupervised baselines on the dialogue topic segmentation task. We can find that ChatGPT performs well in the general domain, exceeding the unsupervised SOTA baseline on almost all datasets. Specifically, ChatGPT achieves the highest performance on DialSeg711, with about 254\% $P_k$ and 115\% $F_1$ scores of SOTA baseline. We attribute this to the clear topic boundaries of DialSeg711, in which topics are usually shifted between booking tickets, hotels, taxis, etc. 
Besides, ChatGPT achieves 179\% $P_k$ and 133\% $F_1$ scores of SOTA baselines on CNTD, but only achieves comparable performance compared with the SOTA baseline on TIAGE. This is because the CNTD dataset is mainly about conversations with background knowledge, making the topics usually focus on a specific argument and thus easy to identify. However, the conversations in the TIAGE dataset are always aimless and without background knowledge. This makes the topics trivial and therefore difficult to recognize. In addition, ChatGPT performs worse than most unsupervised baselines on the ZYS dataset in specific domains. 
%This may be because there are very few domain-specific data are used to train ChatGPT. 
This may be because, in a specific domain, recognizing topic transition requires more domain-specific knowledge as support. 
Therefore, it is difficult for ChatGPT to understand the topics in the specific domains.

% This is mainly because CNTD and TIAGE datasets are about chitchat with and without background knowledge,  mainly focus on fine-grained and coarse-grained topics, respectively. Since ChatGPT tends to give fine-grained topic structures, this makes ChatGPT perform better on CNTD than TIAGE. 

Table~\ref{TS_results_supervised} shows the performance comparison between ChatGPT and supervised baselines. Even under the zero-shot setting, ChatGPT still can achieve 91.33\% and 80.46\% $F_1$ scores of SOTA baseline on CNTD and TIAGE, respectively.
This indicates the great potential of ChatGPT on dialogue topic segmentation.

\subsection{Results on Dialogue Discourse Parsing} Table~\ref{DDP_results} shows the results of ChatGPT on dialogue discourse parsing. ChatGPT achieves 59.91 \emph{Link} $F_1$ and 25.25 \emph{Link\&Rel} $F_1$ scores on STAC, and 63.75 \emph{Link} $F_1$ and 23.85 \emph{Link\&Rel} $F_1$ scores on Molweni. However, the performance of ChatGPT is only comparable to the rule-based method, indicating that ChatGPT only parses rhetorical structure linearly. In addition, ChatGPT achieves about 42\% and 39\% performance of SOTA baseline SDDP in the Link\&Rel metric on the STAC and Molweni, respectively. There is still a large room for improvement as demonstrated by the gap between ChatGPT and the supervised SOTA baselines. It showed a significant challenge ChatGPT faces in understanding hierarchical rhetorical structures.

% On STAC, ChatGPT achieves  on . On Molweni, ChatGPT achieves  on \emph{Link} F1F_1 and  on \emph{Link\&Rel} F1F_1.
% We can observe that ChatGPT can only achieve comparable performance on Link F1F_1 metric with the rule-based method, 

\section{Analysis}
% In this section, we first conducted an ablation study of prompts. Then, we further analyze the abilities of ChatGPT to understand linear and hierarchical structures. Thirdly, we explore the effect of in-context learning on ChatGPT. Then, we analyze the roundness of ChatGPT. Finally, we conducted cases study.

% We replace To explore the effect of different components of prompts, We explored the alternative of each component. specifically, to get the alternative of task description, 
 % To explore the alternative of structured Input, We removed the number of each utterance.

\subsection{ChatGPT's Capability for Discourse Structure}
\begin{figure}[t!]
	\centering  
        \includegraphics[width=\linewidth]{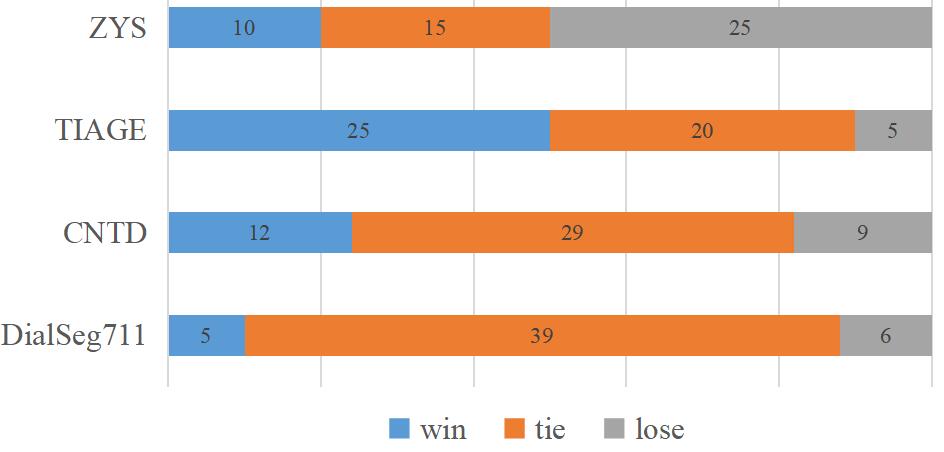}
        \caption{ Manual pair-wise evaluation between ChatGPT-generated and human-annotated topic structures. \textbf{win} indicates that ChatGPT-generated topic structure is more reasonable, \textbf{tie} indicates that ChatGPT-generated and human-annotated topic structures are equally reasonable, and \textbf{lose} indicates that human-annotated topic structure is more reasonable.}
         \label{pair_wise_evaluation}
          \vspace{-0.3cm}
    \end{figure}

\paragraph{ChatGPT understand the linear topic structure of general domain well}
Since different datasets focus on the various granularities of topics, there may be multiple reasonable topic structures for a dialogue. Therefore, solely evaluating the discrepancy between the topic structure predicted by ChatGPT and annotated by humans may underestimate the capability of ChatGPT. 
To further investigate the abilities of ChatGPT to understand topic structures, we randomly selected 50 dialogues from each test set and manually analyzed the ChatGPT-generated and human-annotated topic structures to determine which one is more reasonable. More details are given in Appendix~\ref{detailsofhumanevaluation}. The results are shown in Figure~\ref{pair_wise_evaluation}. On general-domain datasets, i.e., DialSeg711, CNTD, and TIAGE, ChatGPT can provide better or comparable topic structure in more than 80\% of conversations compared with human-annotated. This indicates that ChatGPT can well understand general-domain topics. 

In addition, ChatGPT performs differently in different scenarios of the general domains. Among the three general-domain datasets, ChatGPT performs best in the DialSeg711 dataset. This may be because the dataset is mainly about booking services, with clear topic boundaries, such as the topic shifting from booking a hotel to booking a flight.  Moreover, although CNTD and TIAGE are both chitchat datasets, ChatGPT performs better on CNTD than on TIAGE. This may be because participants in CNTD tend to have conversations around a given reports, leading to the more focused topics and clear topic boundaries. While participants in TIAGE engage in rambling small talk, leading to more trivial and less sustainable topics. 

However, ChatGPT performs poorly in recognizing specific domain topics. On the ZYS dataset, 50\% of the topic structures provided by ChatGPT are inferior to the human-annotated structures. This may be due to ChatGPT having a wealth of general-domain knowledge, but a lack of specialized domain knowledge, such as banking expertise.

\paragraph{ChatGPT hardly understands the hierarchical rhetorical structure}

\begin{figure}[]
	\centering  %
        % \vspace{-0.35cm}
	\subfigbottomskip=2pt %
	\subfigcapskip=-5pt %
	\subfigure{
	\includegraphics[width=0.45\linewidth]{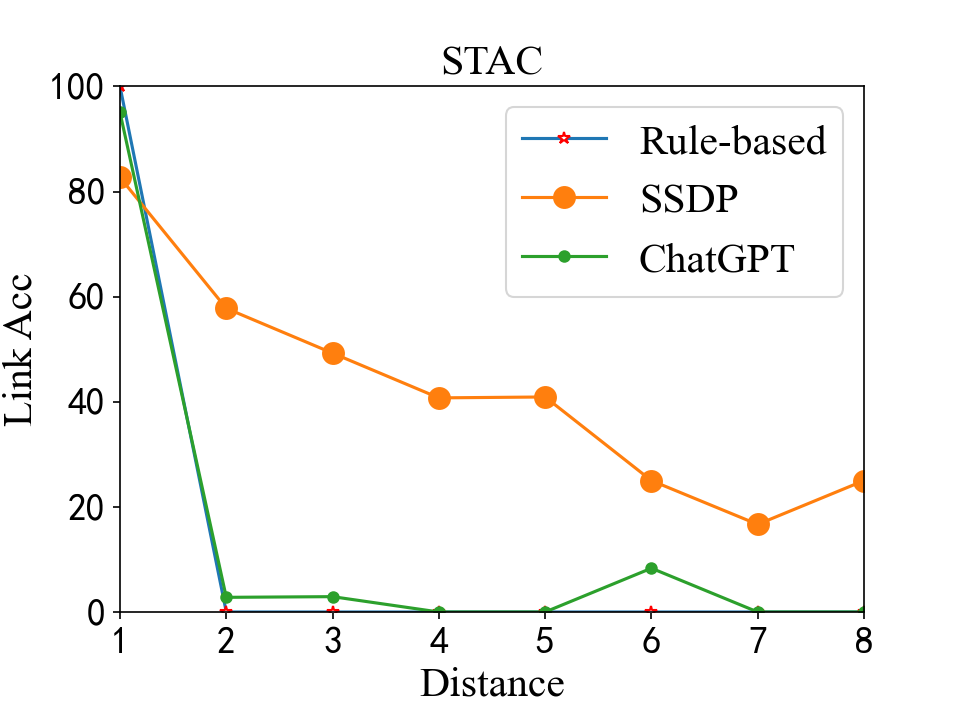}
        % \label{figure4_1}
        }
	\subfigure{
        \includegraphics[width=0.45\linewidth]{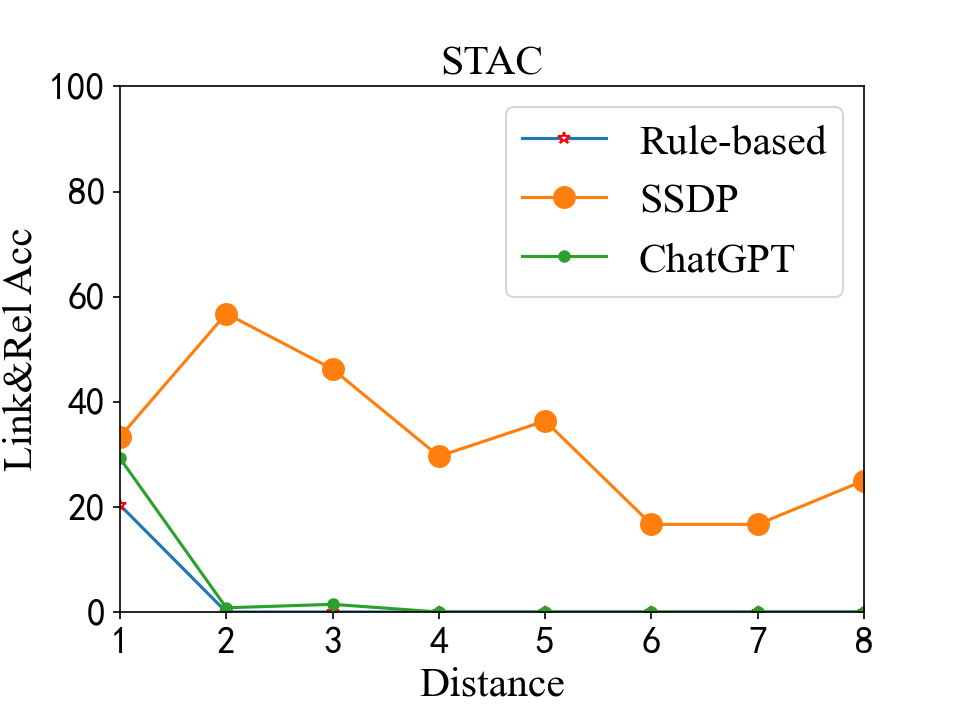}
        % \label{figure4_2}
        }
        \subfigure{
	\includegraphics[width=0.45\linewidth]{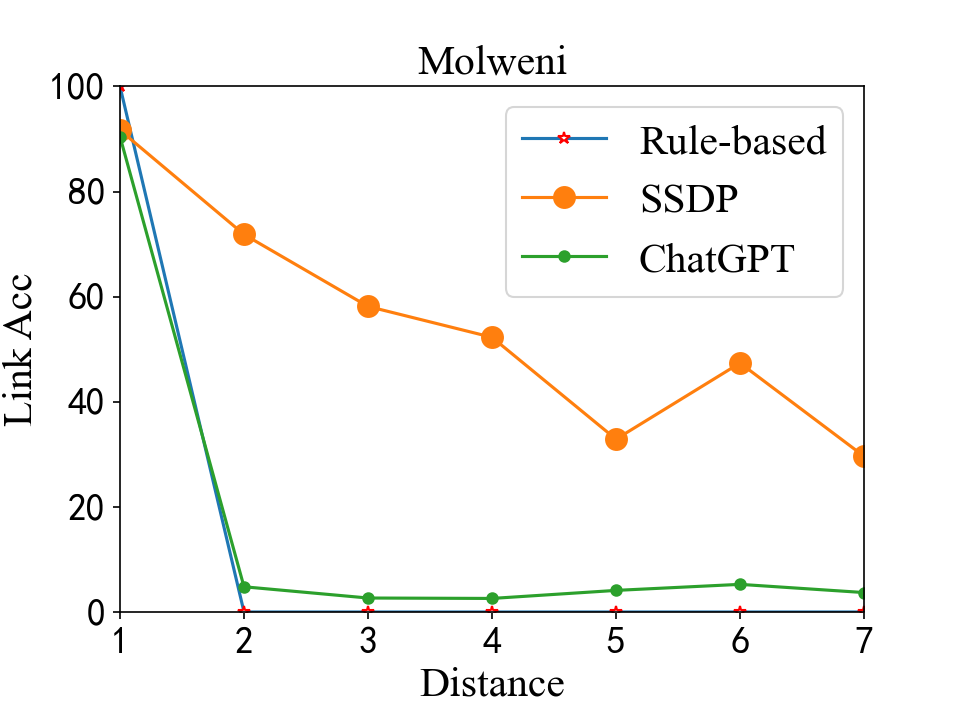}
        % \label{figure4_3}
        }
	\subfigure{
        \includegraphics[width=0.45\linewidth]{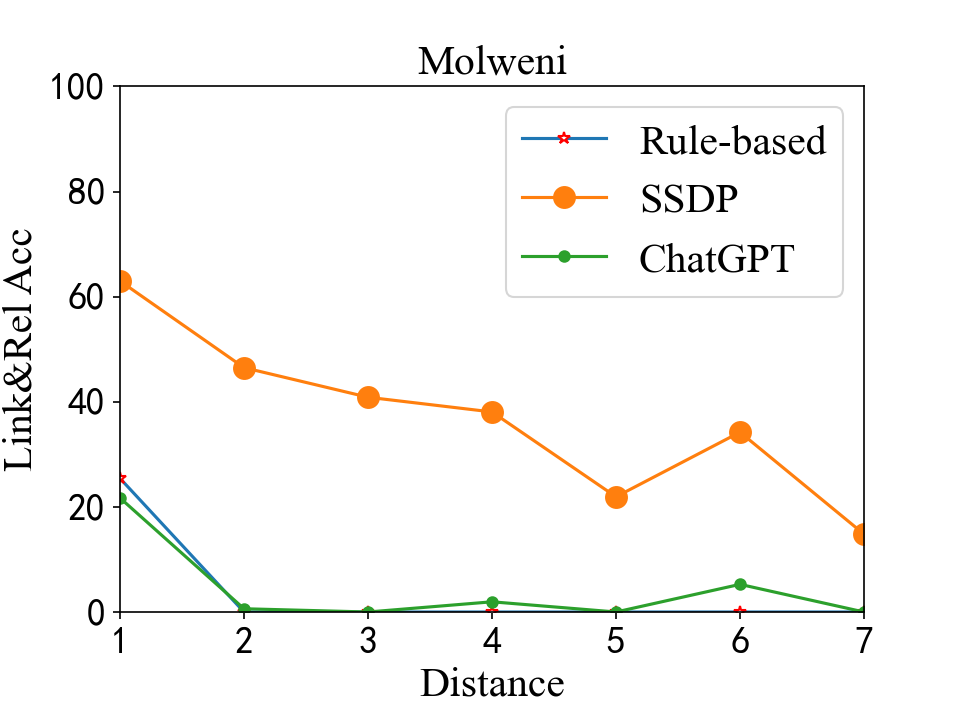}
        % \label{figure4_4}
        }
        \caption{The comparison of performance between ChatGPT and baselines on STAC and Molweni at various distances. If there is a link between $U_j$ and $U_i$, the distance of the link is defined as $i-j$. }
        \label{distance_accuracy}
                      % \vspace{-0.4cm}
    \end{figure}
To further study the abilities of ChatGPT to understand the hierarchical rhetorical structure, we explore the performance on various links with different distances as shown in Figure~\ref{distance_accuracy}. We can observe that the Rule-based method has 100\% link accuracy at the distance 1 on both datasets due to all the links being established in the adjacent utterances. SDDP can recognize the links and relations at various distances and have a downward trend with the distance increasing, which indicates that long-distance links are more difficult to recognize. However, ChatGPT can only recognize the links at a distance of 1 and hardly recognize the links at a distance greater than 1, which has a similar trend with the Rule-based method that parses the hierarchical rhetorical structure linearly. It further suggests that ChatGPT only understands the hierarchical rhetorical structure linearly.

Furthermore, we investigate the performance of ChatGPT in identifying relation types, as shown in Table~\ref{relationperformance}. 
We can observed that ChatGPT mainly recognizes most of the high resource relation types, such as Clarification\_question, Question-answer pair, Acknowledgment, and Continuation.  This suggests that ChatGPT, as a general conversation model, is capable to some extent of understanding the types of relation that are common in the conversations. In addition, even though Comment is also a high resource relation type, it is difficult for ChatGPT to recognize.  This may be because Comment is a type of relation with subjective opinions, while ChatGPT is strictly restricted to output subjective opinions due to its security, which makes it difficult to identify. Finally, for some low resource relation types, such as Correction, Contrast, etc., ChatGPT is usually difficult to identify. This suggests that ChatGPT still faces great challenges in identifying relation in low resource scenarios.

\begin{figure}[t!]
	\centering  %图片全局居中
        \vspace{-0.35cm}
	\subfigbottomskip=2pt %两行子图之间的行间距
	\subfigcapskip=-5pt %设置子图与子标题之间的距离
	\subfigure[]{
	\includegraphics[width=0.8\linewidth]{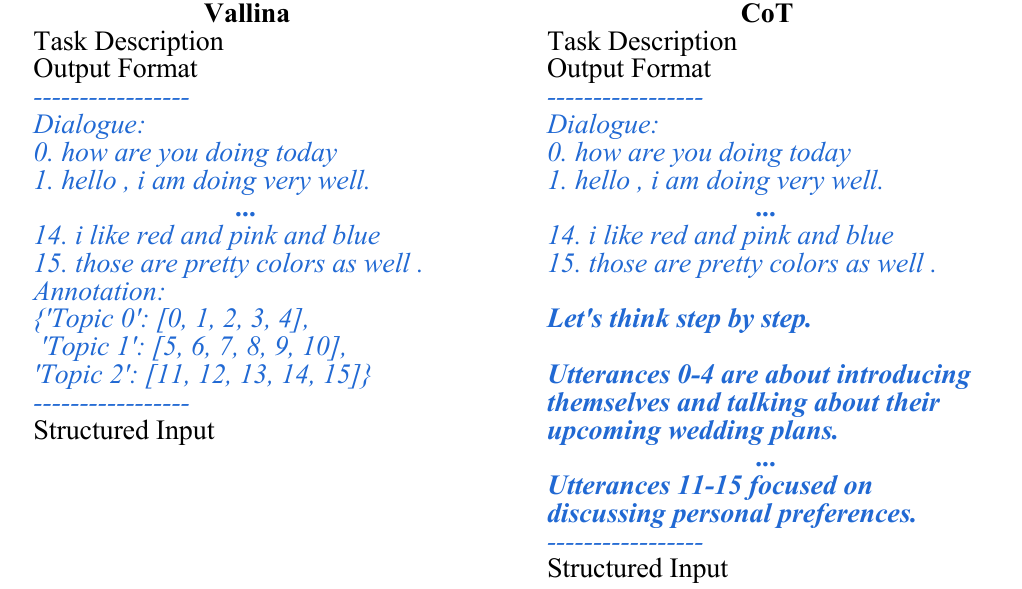}
        % \label{fewshot_1}
        }
        \subfigure[]{
	\includegraphics[width=0.9\linewidth]{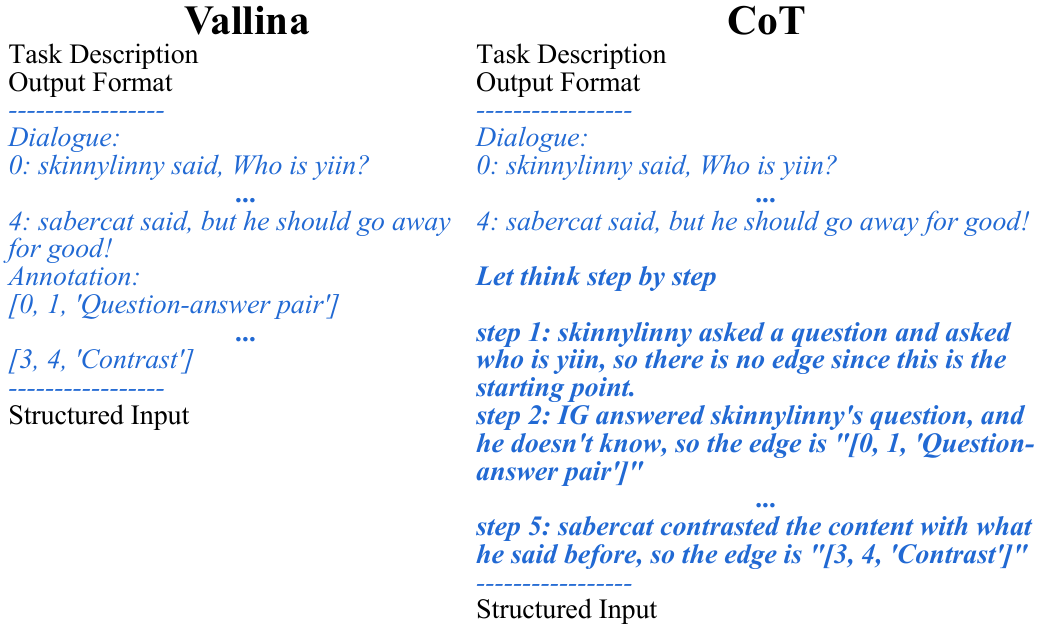}
        % \label{fewshot_3}
        }
        \caption{(a) and (b) show the details of in-context learning with one exemplar for dialogue topic segmentation and dialogue discourse parsing, respectively.}
        \label{few_shot_details}
                      % \vspace{-0.4cm}
    \end{figure}

\begin{figure}[t!]
	\centering
	\subfigure[]{
		\includegraphics[width=0.45\linewidth]{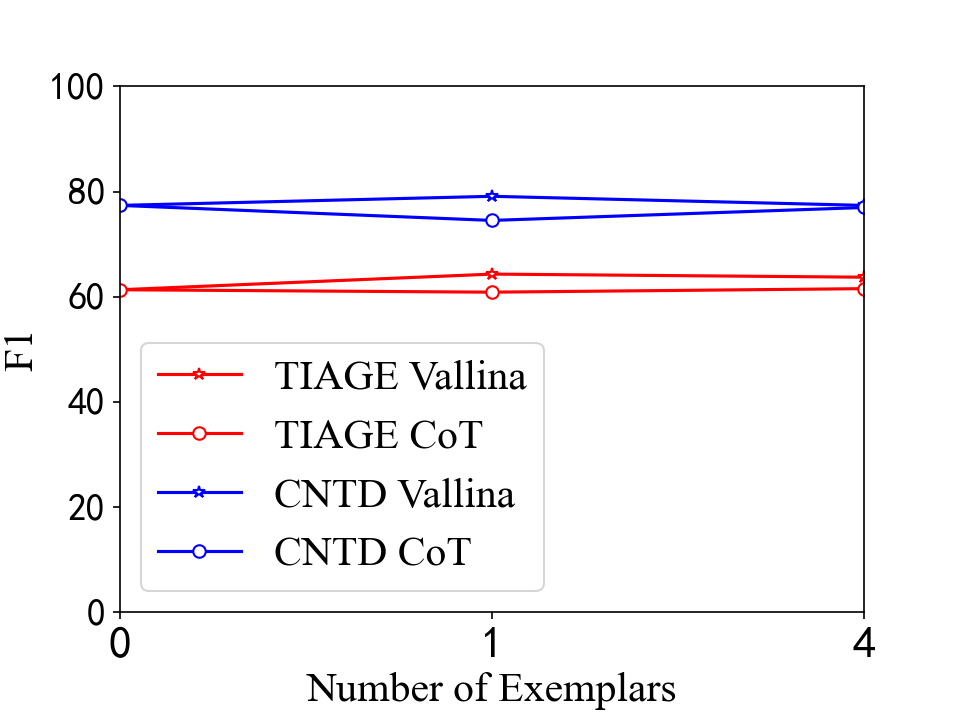}
		\label{fig:subfig1}
	}
	\subfigure[]{
		\includegraphics[width=0.45\linewidth]{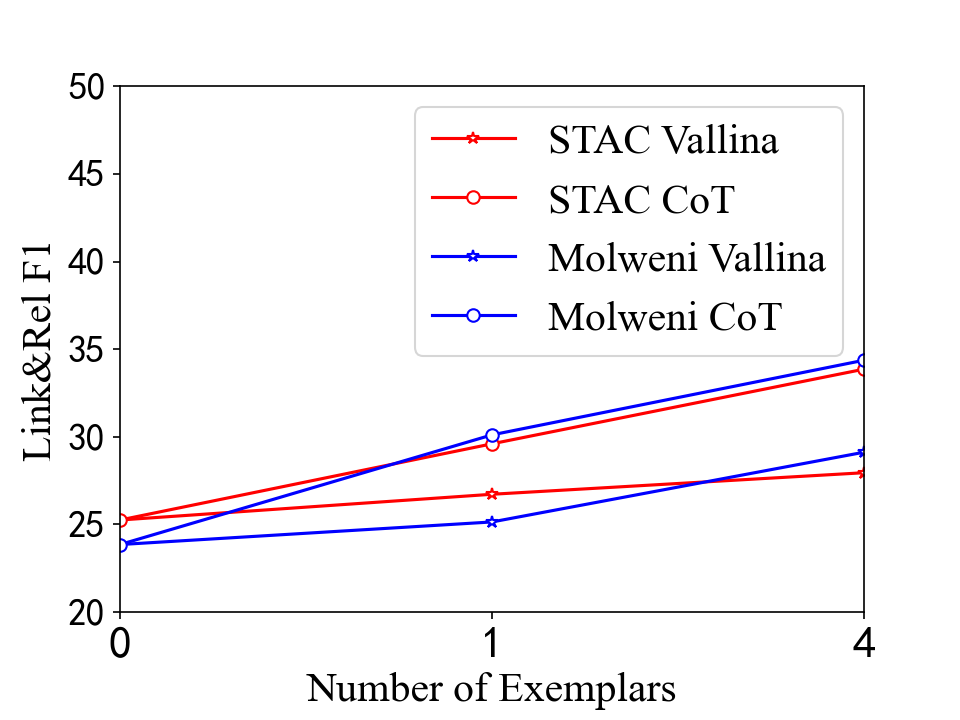}
		\label{fig:subfig2}
	}
	\caption{(a) and (b) show the ICL performance of ChatGPT on dialogue topic segmentation and dialogue discourse parsing, respectively.}
	\label{ICLPerformance}
\end{figure}

% \usepackage{multirow}
\begin{table*}[t!]
\centering
\resizebox{\linewidth}{!}
{ 
\begin{tabular}{ccc|cc}
\hline
\multirow{2}{*}{\textbf{Type}} & \multicolumn{2}{c}{\textbf{Molweni}}           & \multicolumn{2}{c}{\textbf{STAC}}              \\
                      & \textbf{Number(Prop(\%)) }& \textbf{Accuracy(\%)} & \textbf{Number(Prop(\%))} & \textbf{Accuracy(\%)} \\
                      \hline
Comment               & 1081(27.04)            & 1.2          & 165(14.65)             & 7.27         \\
Clarification\_question             & 952(24.34)             & 18.49        & 33(2.93)               & -            \\
Question-answer pair                & 808(20.66)             & 28.84        & 305(27.09)             & 27.21        \\
Continuation          & 275(7.03)              & 10.91        & 113(10.04)             & 19.47        \\
Acknowledgment                  & 131(3.35)              & 11.45        & 147(13.06)             & 18.37        \\
Explanation           & 119(3.04)              & 1.68         & 31(2.75)               & 16.13        \\
Elaboration           & 59(1.51)               & 15.25        & 101(8.97)              & 23.76        \\
Correction            & 53(1.36)               & 1.89         & 21(1.87)               & -            \\
Contrast              & 37(0.95)               & 2.70         & 44(3.81)               & 13.64       \\ \hline
\end{tabular}
}
\caption{Performance of ChatGPT for relation recognition in dialogue discourse parsing task. Prop refers to the proportion of the relation instances.  }
\label{relationperformance}
\end{table*}

\begin{table*}[t!]
\centering
\resizebox{\linewidth}{!}
{ 
\begin{tabular}{ccccc|cc}
\hline
\multicolumn{1}{c}{\multirow{2}{*}{\textbf{Prompt}}} & \multicolumn{4}{c}{\textbf{DTS}}                                     & \multicolumn{2}{c}{\textbf{DDP}} \\
\multicolumn{1}{c}{}                                 & \textbf{DialSeg711} & \textbf{CNTD} & \textbf{TIAGE} & \textbf{ZYS} & \textbf{STAC}     & \textbf{Molweni}     \\ \hline
\multicolumn{1}{c}{Ours}                             & 89.42               & 77.36         & 61.31          & 49.60        & 25.25             & 23.85                \\ \hline
ChatGPT-generated task description                                    & 88.28               & 76.43         & 61.45          & 48.44        & 23.82             & 23.26                \\
Sequence labeling/Adjacency matrix output                                        & 51.95               & 59.05         & 53.27          & 40.67        & 12.53             & 13.37                \\
Unstructured input                                      & 83.21               & 74.01         & 58.69          & 44.62        & 21.54             & 21.22       \\ \hline        
\end{tabular}
}
\caption{Ablation study of different components of prompt. $F_1$ and \emph{Link\&Rel} $F_1$ performance are reported for Dialogue Topic Segmentation (DTS) and Dialogue Discourse Parsing (DDP), respectively. }
\label{prompts_ablation}

\end{table*}

\begin{table}[t!]
\centering

\setlength{\tabcolsep}{0.8mm}{ 

\begin{tabular}{ccccc}
 \hline 
Task                                         & Dataset    & \#Dial. & Avg.\#NM. & Ratio.(\%) \\ \hline
\multirow{4}{*}{DTS} & TIAGE      & 100     & 0.8            &0.80       \\
                                             & CNTD       & 133     & 3.2            & 2.40       \\
                                             & DialSeg711 & 711     & 68             & 9.56      \\
                                             & ZYS        & 505     & 24             & 4.75       \\ \hline
\multirow{2}{*}{DDP}  & STAC       & 111     & 2              & 1.80       \\
                                             & Molweni    & 500     & 3.6              & 0.72 \\ \hline    
\end{tabular}
}
\caption{The ratio that ChatGPT does not follow the output format on each dataset. “\#Dial.” is the number of dialogue for testing. “Avg. \#NM.” indicates the average number of dialogues that ChatGPT does not follow the output format under two-times experiments. “Ratio (\%)” denotes the percentage of “Avg. \#NM.” and “\#Dial.”.}
\label{notmeet}
\end{table}

\subsection{Impact of In Context Learning on ChatGPT}
In this section, we investigated the impact of in-context learning~\cite{min-etal-2022-rethinking} on ChatGPT's understanding of discourse structure.
We select at most 4 exemplars randomly from the training set of the supervised dataset due to the token limitation of ChatGPT. Then, we introduce the exemplars into the prompt in two types as shown in Figure~\ref{few_shot_details}: Vanilla and Chain-of-Thought (CoT)\cite{jason_wei_chain}. The vanilla type directly inserts the dialogue and its discourse structure between the output format and structured input in the prompt. The CoT type not only provides the dialogue and its discourse structure but also some intermediate steps are written manually for deriving discourse structure. We conduct the experiments two times and report the average performance.

Figure~\ref{ICLPerformance} shows that neither Vanilla nor CoT few-shot can improve the performance of ChatGPT on dialogue topic segmentation with the number of exemplars increasing. This shows that ChatGPT has a good understanding of linear topic structures and more exemplars can not enhance the capabilities of ChatGPT on topic structure significantly. 

However, both vanilla and CoT types can improve the performance of ChatGPT on dialogue discourse parsing as the number of exemplars increases. This demonstrates the insufficiency of ChatGPT to understand hierarchical rhetorical structures and more exemplars can significantly alleviate the deficiencies of ChatGPT.
It is worth noting that the CoT type helps ChatGPT more since it provides intermediate steps for deriving discourse structures. This suggests that dialogue discourse parsing is a complex task that requires multi-step reasoning.

\subsection{Ablation Study of Prompt Components}
To study the impact of the three components (task description, output format, and structured input) that make up the prompt on the performance of two tasks, we conducted an ablation study as shown in Table~\ref{prompts_ablation}. More details about the variants of different components are given in Appendix~\ref{detailsofvariants}.

We first investigated the impact of different sources of task descriptions (manually written and ChatGPT generated) on performance. Specifically, we used the original manually written task descriptions as a reference to ask ChatGPT to generate corresponding task descriptions. The results shown in the third row of Table~\ref{prompts_ablation} indicate that the performance difference between the task descriptions generated by ChatGPT and those written manually is not significant, suggesting that the performance of ChatGPT is not sensitive to changes in task descriptions.

Then, we delved into the impact of different output forms (ours and traditional) on performance. Specifically, for dialogue topic segmentation, we instruct ChatGPT to output a 0/1 sequence for the utterance list, where '1' indicates the current utterance is the end of a topic, as shown in Figure~\ref{description_task}. For dialogue discourse parsing, we replace the sparse matrix we adopted with an adjacency matrix. The results in the fourth row of Table~\ref{prompts_ablation} showed a significant performance degradation on both tasks. We found it is because ChatGPT may be insensitive to the number of utterances in a dialogue by our error analysis. For example, for a dialogue with $N$ utterances, ChatGPT cannot always output a 0/1 list of length $N$ to represent the topic structure, or an adjacency matrix of shape $N \times N$ to represent the discourse structure, with a percentage of 52\% and 65\%, respectively,  which introduces serious evaluation errors. 

Finally, we explored the impact of structured or unstructured inputs on performance. The last row of Table~\ref{prompts_ablation} showed there is only slight performance degradation after removing the structured number. Error analysis reveals that ChatGPT ignores some of the utterances without the help of the number indication, which requires more post-processing operations for evaluation, resulting in performance degradation.

\begin{table*}[t] \scriptsize
\centering
\begin{tabular}{cccc}
\hline
\textbf{Number} &  \textbf{Utterance} &  \textbf{Annotation} &  \textbf{ChatGPT} \\
\hline
 0  &  how are you ? being an old man , i am   slowing down these days                  &  0         &  0       \\
 1  &  hi , my dad is old as well , they live   close to me and i see them often       &  0         &  0       \\
 2  &  that is a great thing honor your dad   with your presence                        &  0         &  0      \\
 3  &  sure , i pick him up for church every   sunday with my ford pickup               &  1         &  1       \\
 4 &  sounds wonderful my wheelchair can go   very fast on various terrains            &  0         &  0       \\
 5  &  i guess that means you do not go   hunting often ? i love hunting , i own 3 guns &  0         &  0       \\
 6  &  hunting ? i served in the marines , yes   i hunt                                 &  0         &  0       \\
 7  &  yeah me too , i am conservative so i   love church and hunting                  &  0        &  1       \\
 8  &  what do you like to hunt ? do you ever   fish ?                                  &  0         &  0       \\
 9 &  fishing is good . i love fishing as   well                                      &  0         &  0       \\
 10 &  fishing is a better choice sometimes   for my one leg                            &  0         &  0       \\
 11 &  yes that must be hard , i hope things   get better for you                       &  0         &  0       \\
 12 &  i enjoy life , it is what it is these   days .                                  &  0         &  0       \\
 13 &  yes i agree . i try to enjoy life too ,   whenever i am not working              &  0         &  1       \\
 14 &  well , you better enjoying working so   you can enjoy more of your time .        &  0         &  0       \\
 15 &  yeah but i go to church every sunday so   my weekends are usually booked         &  0         &  0      \\ 
 \hline
\end{tabular}
\caption{Case study for dialogue topic segmentation and the cases are from the TIAGE dataset. '1' indicates the end of a topic.}

\label{case_DTS}
\end{table*}

\begin{table*}[t] \scriptsize
\centering
\begin{tabular}{ccccc}
\hline
\multicolumn{1}{c}{\textbf{Number}} & \multicolumn{1}{c}{\textbf{Speaker}} & \multicolumn{1}{c}{\textbf{Utterance}}                   & \multicolumn{1}{c}{\textbf{Annotation}}                                                                                                                                                                                                                                                                            & \multicolumn{1}{c}{ChatGPT}                                                                                                           \\
0                      & somdechn                    & I need wood, clay or ore,I can give Sheep       & \multirow{11}{*}{\begin{tabular}[c]{@{}l@{}}(0,1): Question-answer\_pair\\ (0,2): Question-answer\_pair\\ (2,3): Comment\\ (1,4): Question-Elaboration\\ (4,5): Question-answer\_pair\\ (2,6): Elaboration\\ (6,7): Explanation\\ (5,8): Comment\\ (8,9): Continuation\\ (9,10): Correction\end{tabular}} & \multirow{11}{*}{\begin{tabular}[c]{@{}l@{}}(0,1): Elaboration\\ (1,2): Comment\\ (2,3): Comment\\ (3,4): Continuation\\ (4,5): Question-answer\_pair\\ (5,6): Comment\\ (6,7): Continuation\\ (7,8): Comment\\ (8,9): Continuation\\ (9,10): Correction\end{tabular}} \\
1                      & Shawnus                     & i can trade wood                                &                                                                                                                                                                                                                                                                                                          &                                                                                                                                                                                                                                                                        \\
2                      & ztime                       & just spent it all                               &                                                                                                                                                                                                                                                                                                          &                                                                                                                                                                                                                                                                        \\
3                      & ztime                       & sorry                                           &                                                                                                                                                                                                                                                                                                          &                                                                                                                                                                                                                                                                        \\
4                      & somdechn                    & 1 sheep for 1 wood?                             &                                                                                                                                                                                                                                                                                                          &                                                                                                                                                                                                                                                                        \\
5                      & Shawnus                     & 2 sheep 1 wood                                  &                                                                                                                                                                                                                                                                                                          &                                                                                                                                                                                                                                                                        \\
6                      & ztime                       & sorry empty                                     &                                                                                                                                                                                                                                                                                                          &                                                                                                                                                                                                                                                                        \\
7                      & ztime                       & tough times..                                   &                                                                                                                                                                                                                                                                                                          &                                                                                                                                                                                                                                                                        \\
8                      & Shawnus                     & hopefully i dont roll a 7                       &                                                                                                                                                                                                                                                                                                          &                                                                                                                                                                                                                                                                        \\
9                      & Shawnus                     & and that biotes me in the arse                  &                                                                                                                                                                                                                                                                                                          &                                                                                                                                                                                                                                                                        \\
10                     & Shawnus                     & bites*                                          &                                                                                                                                                                                                                                                                                                          &                                \\ \hline                                                                                                                                                                                                                                        
\end{tabular}

\caption{Case study for dialogue discourse parsing and the case are from STAC dataset.}
\label{case_DDP}
\end{table*}

\subsection{Robustness of ChatGPT}
Because ChatGPT can not always output in the specified format, we investigate the instruction-following capabilities of ChatGPT. For each dataset, we report the ratio that ChatGPT does not follow the output format as shown in Table~\ref{notmeet}. We can see that ChatGPT can not fully follow the instructions on each dataset, reaching a maximum of 9.56\% on the DialSeg711. We analyzed the samples on DialSeg711 that do not follow instructions and found that more than 90\% required inter-topic processing, as shown in Figure~\ref{post_processing_figure}. We attribute this to the fact that ChatGPT may ignore topics containing fewer utterances during the generation even if it has recognized the topics. This indicates that ChatGPT's robustness is still an issue for task completion.

\subsection{Case Study}
To further demonstrate ChatGPT's success in understanding linear topic structure and failure in hierarchical rhetorical structure, we conducted case studies as shown in Table~\ref{case_DTS} and Table~\ref{case_DDP}.
From the case in Table~\ref{case_DTS}, we can see that human annotation only divided the dialogue into two topics ( $U_1$-$U_4$ and $U_5$-$U_{16}$). However, there are more topics among $U_5$ to $U_{16}$, including hunting, fishing, and enjoying life.  ChatGPT successfully identifies these topics, giving reasonable topic boundaries. This indicates that ChatGPT can understand the linear topic structure well in the general domain.

Table~\ref{case_DDP} shows a case from the STAC dataset. Human annotation annotates several discourse relations between utterances with longer distances, such as \emph{(1, 4): Question-Elaboration}, \emph{(2, 6): Elaboration}, etc. However, ChatGPT always establishes the discourse relation between adjacent utterances linearly, showing it hardly understands the hierarchical rhetorical structure.

% \subsection{Capabilities of Open-source LLMs for Discourse Analysis in Dialogue }
\section{Conclusion}
In this paper, we conducted a systematic inspection of ChatGPT’s capabilities in two dialogue discourse tasks (topic segmentation and discourse parsing) for its deep semantic understanding of  linear and hierarchical discourse structures. We first crafted the prompt template with the task description, output format, and structured input to guide ChatGPT to complete the task. Then, we conducted the experiments on four popular topic segmentation datasets and two discourse parsing datasets. The experimental results reveal that ChatGPT has a good understanding of topic structure in general-domain conversations but struggles in specific-domain conversations. Besides, ChatGPT hardly understands the rhetorical structure, which is more complex and needs to consider long-distant dependent relations of utterances. In-depth analysis indicates that ChatGPT could give finer-granularity topic structures than human annotations but only parses the hierarchical rhetorical structures linearly. Besides, we delved into the impact of in-context learning on ChatGPT and observed that chain of thought can significantly improve the capabilities of ChatGPT to parse the hierarchical structures. In addition, we delved into the impact of various prompt components and observed that output format contributes the most. Finally, the robustness of ChatGPT is still an issue of concern. We hope these findings provide a foundation for dialogue discourse analysis in future research.

%In this paper, we conducted a systematic inspection of ChatGPT’s performance in two dialogue discourse tasks: topic segmentation and discourse parsing, for ChatGPT’s ability to understand linear topics and hierarchical rhetorical structures. We first crafted the prompt consisting of the task description, output format, and structured input, to guide ChatGPT to complete the task. Then, we conducted the experiments on four popular topic segmentation datasets and two discourse parsing datasets. The experimental results reveal that ChatGPT has a good understanding of general-domain topics but struggles in identifying specific-domain topics. Besides, ChatGPT can hardly understand the rhetorical structures. In-depth analysis indicates that ChatGPT can give more reasonable topic structures than human annotations but only parses the hierarchical rhetorical structures linearly. Besides, we delved into the impact of in-context learning on ChatGPT and observed that chain of thought can significantly improve the capabilities of ChatGPT to parse the hierarchical structures. Furthermore, we conducted the ablation study on various prompt components and found that the output format contributes the most. In addition, the robustness of ChatGPT is still an issue of concern. We hope these findings provide a foundation for dialogue discourse analysis in future research.

\section{Acknowledgements}
%Place all acknowledgments (including those concerning research grants and funding) in a separate section at the end of the paper.
The authors would like to thank the three anonymous reviewers for their comments on this paper.
This research is supported by the National Natural Science Foundation of China (No. 62376181),  the Key R\&D Plan of Jiangsu Province (BE2021048), the project of Shenzhen Science and Technology Research Fund (Fundamental Research Key Project Grant No. JCYJ20220818103001002), the Internal Project Fund from Shenzhen Research Institute of Big Data under Grant No. T00120-220002, and Shenzhen Key Laboratory of Cross-Modal Cognitive Computing (grant number ZDSYS20230626091302006).

% \section*{Limitations}
% The main limitation of our work is the selection of prompts. The prompts are subject to human understanding and the experimental cost prohibits us from conducting more experiments on the selection of prompts. Hence, we cannot guarantee whether other prompts can help ChatGPT achieve better performance. Furthermore, due to the continuous version updates and iterations of ChatGPT, it is hard to guarantee that future versions can lead to similar results to ours.

\nocite{*}
\section{Bibliographical References}\label{sec:reference}

\bibliographystyle{lrec-coling2024-natbib}
\bibliography{lrec-coling2024-example}

\begin{thebibliography}{54}
\expandafter\ifx\csname natexlab\endcsname\relax\def\natexlab#1{#1}\fi

\bibitem[{Asher et~al.(2016)Asher, Hunter, Morey, Benamara, and Afantenos}]{asher2016discourse}
Nicholas Asher, Julie Hunter, Mathieu Morey, Farah Benamara, and Stergos Afantenos. 2016.
\newblock {Discourse Structure and Dialogue Acts in Multiparty Dialogue: the STAC Corpus}.
\newblock In \emph{Proceedings of the Tenth International Conference on Language Resources and Evaluation}, pages 2721--2727.

\bibitem[{Asher and Lascarides(2003)}]{asher2003logics}
Nicholas Asher and Alex Lascarides. 2003.
\newblock \emph{Logics of conversation}.
\newblock Cambridge University Press.

\bibitem[{Beeferman et~al.(1999)Beeferman, Berger, and Lafferty}]{beeferman1999statistical}
Doug Beeferman, Adam Berger, and John Lafferty. 1999.
\newblock {Statistical Models for Text Segmentation}.
\newblock \emph{Machine learning}, 34:177--210.

\bibitem[{Budzianowski et~al.(2018)Budzianowski, Wen, Tseng, Casanueva, Ultes, Ramadan, and Ga{\v{s}}i{\'c}}]{budzianowski-etal-2018-multiwoz}
Pawe{\l} Budzianowski, Tsung-Hsien Wen, Bo-Hsiang Tseng, I{\~n}igo Casanueva, Stefan Ultes, Osman Ramadan, and Milica Ga{\v{s}}i{\'c}. 2018.
\newblock {{M}ulti{WOZ} - A Large-Scale Multi-Domain {W}izard-of-{O}z Dataset for Task-Oriented Dialogue Modelling}.
\newblock In \emph{Proceedings of the 2018 Conference on Empirical Methods in Natural Language Processing}, pages 5016--5026.

\bibitem[{Chan et~al.(2023)Chan, Cheng, Wang, Jiang, Fang, Liu, and Song}]{chan2023chatgpt}
Chunkit Chan, Jiayang Cheng, Weiqi Wang, Yuxin Jiang, Tianqing Fang, Xin Liu, and Yangqiu Song. 2023.
\newblock {ChatGPT Evaluation on Sentence Level Relations: A Focus on Temporal, Causal, and Discourse Relations}.
\newblock \emph{arXiv preprint arXiv: 2304.14827}.

\bibitem[{Chi and Rudnicky(2022)}]{chi-rudnicky-2022-structured}
Ta-Chung Chi and Alexander Rudnicky. 2022.
\newblock {Structured Dialogue Discourse Parsing}.
\newblock In \emph{Proceedings of the 23rd Annual Meeting of the Special Interest Group on Discourse and Dialogue}, pages 325--335.

\bibitem[{Devlin et~al.(2019)Devlin, Chang, Lee, and Toutanova}]{devlin2019bert}
Jacob Devlin, Ming-Wei Chang, Kenton Lee, and Kristina Toutanova. 2019.
\newblock {BERT: Pre-training of Deep Bidirectional Transformers for Language Understanding}.
\newblock In \emph{Proceedings of the 2019 Conference of the North {A}merican Chapter of the Association for Computational Linguistics: Human Language Technologies}, pages 4171--4186.

\bibitem[{Eric et~al.(2017)Eric, Krishnan, Charette, and Manning}]{eric-etal-2017-key}
Mihail Eric, Lakshmi Krishnan, Francois Charette, and Christopher~D. Manning. 2017.
\newblock {Key-Value Retrieval Networks for Task-Oriented Dialogue}.
\newblock In \emph{Proceedings of the 18th Annual {SIGDIAL} Meeting on Discourse and Dialogue}, pages 37--49.

\bibitem[{Fan et~al.(2023)Fan, Jiang, Li, Kong, and Zhu}]{fan-etal-2023-improving}
Yaxin Fan, Feng Jiang, Peifeng Li, Fang Kong, and Qiaoming Zhu. 2023.
\newblock Improving dialogue discourse parsing via reply-to structures of addressee recognition.
\newblock In \emph{Proceedings of the 2023 Conference on Empirical Methods in Natural Language Processing}, pages 8484--8495.

\bibitem[{Fan et~al.(2022)Fan, Li, Kong, and Zhu}]{fan2022distance}
Yaxin Fan, Peifeng Li, Fang Kong, and Qiaoming Zhu. 2022.
\newblock {A Distance-Aware Multi-Task Framework for Conversational Discourse Parsing}.
\newblock In \emph{Proceedings of the 29th International Conference on Computational Linguistics}, pages 912--921.

\bibitem[{Gao et~al.(2023{\natexlab{a}})Gao, Wang, Lin, Wu, Yang, Huang, and Li}]{gao2023unsupervised}
Haoyu Gao, Rui Wang, Ting-En Lin, Yuchuan Wu, Min Yang, Fei Huang, and Yongbin Li. 2023{\natexlab{a}}.
\newblock {Unsupervised Dialogue Topic Segmentation with Topic-aware Contrastive Learning}.
\newblock In \emph{Proceedings of the 46th International ACM SIGIR Conference on Research and Development in Information Retrieval}, pages 2481--2485.

\bibitem[{Gao et~al.(2023{\natexlab{b}})Gao, Zhao, Yu, and Xu}]{gao2023exploring}
Jun Gao, Huan Zhao, Changlong Yu, and Ruifeng Xu. 2023{\natexlab{b}}.
\newblock {Exploring the Feasibility of ChatGPT for Event Extraction}.
\newblock \emph{arXiv preprint arXiv: 2303.03836}.

\bibitem[{Han et~al.(2023)Han, Peng, Yang, Wang, Liu, and Wan}]{han2023information}
Ridong Han, Tao Peng, Chaohao Yang, Benyou Wang, Lu~Liu, and Xiang Wan. 2023.
\newblock {Is Information Extraction Solved by ChatGPT? An Analysis of Performance, Evaluation Criteria, Robustness and Errors}.
\newblock \emph{arXiv preprint arXiv: 2305.14450}.

\bibitem[{He et~al.(2021)He, Zhang, and Zhao}]{he2021multi}
Yuchen He, Zhuosheng Zhang, and Hai Zhao. 2021.
\newblock {Multi-tasking Dialogue Comprehension with Discourse Parsing}.
\newblock In \emph{Proceedings of the 35th Pacific Asia Conference on Language, Information and Computation}, pages 598--608.

\bibitem[{Hearst(1997)}]{hearst1997text}
Marti~A Hearst. 1997.
\newblock {TextTiling: Segmenting Text into Multi-paragraph Subtopic Passages}.
\newblock \emph{Computational linguistics}, 23(1):33--64.

\bibitem[{Hu et~al.(2023)Hu, Ameer, Zuo, Peng, Zhou, Li, Li, Li, Jiang, and Xu}]{hu2023zeroshot}
Yan Hu, Iqra Ameer, Xu~Zuo, Xueqing Peng, Yujia Zhou, Zehan Li, Yiming Li, Jianfu Li, Xiaoqian Jiang, and Hua Xu. 2023.
\newblock {Zero-shot Clinical Entity Recognition using ChatGPT}.
\newblock \emph{arXiv preprint arXiv: 2303.16416}.

\bibitem[{Jiao et~al.(2023)Jiao, Wang, tse Huang, Wang, and Tu}]{jiao2023chatgpt}
Wenxiang Jiao, Wenxuan Wang, Jen tse Huang, Xing Wang, and Zhaopeng Tu. 2023.
\newblock {Is ChatGPT A Good Translator? Yes With GPT-4 As The Engine}.
\newblock \emph{arXiv preprint arXiv: 2301.08745}.

\bibitem[{Li et~al.(2020)Li, Liu, Kan, Zheng, Wang, Lei, Liu, and Qin}]{li-etal-2020-molweni}
Jiaqi Li, Ming Liu, Min-Yen Kan, Zihao Zheng, Zekun Wang, Wenqiang Lei, Ting Liu, and Bing Qin. 2020.
\newblock {Molweni: A Challenge Multiparty Dialogues-based Machine Reading Comprehension Dataset with Discourse Structure}.
\newblock In \emph{Proceedings of the 28th International Conference on Computational Linguistics}, pages 2642--2652.

\bibitem[{Li et~al.(2017)Li, Su, Shen, Li, Cao, and Niu}]{li-etal-2017-dailydialog}
Yanran Li, Hui Su, Xiaoyu Shen, Wenjie Li, Ziqiang Cao, and Shuzi Niu. 2017.
\newblock {{D}aily{D}ialog: A Manually Labelled Multi-turn Dialogue Dataset}.
\newblock In \emph{Proceedings of the Eighth International Joint Conference on Natural Language Processing}, pages 986--995.

\bibitem[{Liang et~al.(2023)Liang, Wu, Cui, Bai, Bian, and Li}]{liang2023enhancing}
Xinnian Liang, Shuangzhi Wu, Chenhao Cui, Jiaqi Bai, Chao Bian, and Zhoujun Li. 2023.
\newblock Enhancing dialogue summarization with topic-aware global- and local- level centrality.
\newblock In \emph{Proceedings of the 17th Conference of the European Chapter of the Association for Computational Linguistics}, pages 27--38.

\bibitem[{Lin et~al.(2023{\natexlab{a}})Lin, Fan, Chu, Li, and Zhu}]{lin2023multigranularity}
Jiangyi Lin, Yaxin Fan, Xiaomin Chu, Peifeng Li, and Qiaoming Zhu. 2023{\natexlab{a}}.
\newblock {Multi-Granularity Prompts for Topic Shift Detection in Dialogue}.
\newblock In \emph{Proceedings of the 2023 Nineteenth International Conference on Intelligent Computing}, pages 511--522.

\bibitem[{Lin et~al.(2023{\natexlab{b}})Lin, Fan, Jiang, Chu, and Li}]{lin2023topic}
Jiangyi Lin, Yaxin Fan, Feng Jiang, Xiaomin Chu, and Peifeng Li. 2023{\natexlab{b}}.
\newblock {Topic Shift Detection in Chinese Dialogues: Corpus and Benchmark}.
\newblock In \emph{Proceedings of the 17th International Conference on Document Analysis and Recognition}, pages 166--183.

\bibitem[{Liu and Chen(2021)}]{liu2021improving}
Zhengyuan Liu and Nancy Chen. 2021.
\newblock {Improving Multi-Party Dialogue Discourse Parsing via Domain Integration}.
\newblock In \emph{Proceedings of the 2nd Workshop on Computational Approaches to Discourse}, pages 122--127.

\bibitem[{Lowe et~al.(2015)Lowe, Pow, Serban, and Pineau}]{lowe2015ubuntu}
Ryan Lowe, Nissan Pow, Iulian~Vlad Serban, and Joelle Pineau. 2015.
\newblock {The Ubuntu Dialogue Corpus: A Large Dataset for Research in Unstructured Multi-Turn Dialogue Systems}.
\newblock In \emph{Proceedings of the 16th Annual Meeting of the Special Interest Group on Discourse and Dialogue}, pages 285--294.

\bibitem[{Min et~al.(2022)Min, Lyu, Holtzman, Artetxe, Lewis, Hajishirzi, and Zettlemoyer}]{min-etal-2022-rethinking}
Sewon Min, Xinxi Lyu, Ari Holtzman, Mikel Artetxe, Mike Lewis, Hannaneh Hajishirzi, and Luke Zettlemoyer. 2022.
\newblock {Rethinking the Role of Demonstrations: What Makes In-Context Learning Work?}
\newblock In \emph{Proceedings of the 2022 Conference on Empirical Methods in Natural Language Processing}, pages 11048--11064.

\bibitem[{Nori et~al.(2023)Nori, King, McKinney, Carignan, and Horvitz}]{nori2023capabilities}
Harsha Nori, Nicholas King, Scott~Mayer McKinney, Dean Carignan, and Eric Horvitz. 2023.
\newblock {Capabilities of GPT-4 on Medical Challenge Problems}.
\newblock \emph{arXiv preprint arXiv: 2303.13375}.

\bibitem[{Pevzner and Hearst(2002)}]{pevzner2002critique}
Lev Pevzner and Marti~A. Hearst. 2002.
\newblock {A Critique and Improvement of an Evaluation Metric for Text Segmentation}.
\newblock \emph{Computational Linguistics}, 28(1):19--36.

\bibitem[{Pu and Demberg(2023)}]{pu2023chatgpt}
Dongqi Pu and Vera Demberg. 2023.
\newblock {ChatGPT vs Human-authored Text: Insights into Controllable Text Summarization and Sentence Style Transfer}.
\newblock \emph{arXiv preprint arXiv: 2306.07799}.

\bibitem[{Qin et~al.(2023)Qin, Zhang, Zhang, Chen, Yasunaga, and Yang}]{qin2023chatgpt}
Chengwei Qin, Aston Zhang, Zhuosheng Zhang, Jiaao Chen, Michihiro Yasunaga, and Diyi Yang. 2023.
\newblock {Is ChatGPT a General-Purpose Natural Language Processing Task Solver?}
\newblock \emph{arXiv preprint arXiv: 2302.06476}.

\bibitem[{Raffel et~al.(2020)Raffel, Shazeer, Roberts, Lee, Narang, Matena, Zhou, Li, and Liu}]{raffel2020exploring}
Colin Raffel, Noam Shazeer, Adam Roberts, Katherine Lee, Sharan Narang, Michael Matena, Yanqi Zhou, Wei Li, and Peter~J. Liu. 2020.
\newblock {Exploring the Limits of Transfer Learning with a Unified Text-to-Text Transformer}.
\newblock \emph{arXiv preprint arXiv: 1910.10683}.

\bibitem[{Robinson et~al.(2023)Robinson, Rytting, and Wingate}]{robinson2023leveraging}
Joshua Robinson, Christopher~Michael Rytting, and David Wingate. 2023.
\newblock {Leveraging Large Language Models for Multiple Choice Question Answering}.
\newblock \emph{arXiv preprint arXiv: 2210.12353}.

\bibitem[{Shi and Huang(2019)}]{shi2019deep}
Zhouxing Shi and Minlie Huang. 2019.
\newblock {A Deep Sequential Model for Discourse Parsing on Multi-Party Dialogues}.
\newblock In \emph{Proceedings of the AAAI Conference on Artificial Intelligence}, pages 7007--7014.

\bibitem[{Solbiati et~al.(2021)Solbiati, Heffernan, Damaskinos, Poddar, Modi, and Cali}]{solbiati2021unsupervised}
Alessandro Solbiati, Kevin Heffernan, Georgios Damaskinos, Shivani Poddar, Shubham Modi, and Jacques Cali. 2021.
\newblock {Unsupervised Topic Segmentation of Meetings with BERT Embeddings}.
\newblock \emph{arXiv preprint arXiv: 2106.12978}.

\bibitem[{Song et~al.(2016)Song, Mou, Yan, Yi, Zhu, Hu, and Zhang}]{song2016dialogue}
Yiping Song, Lili Mou, Rui Yan, Li~Yi, Zinan Zhu, Xiaohua Hu, and Ming Zhang. 2016.
\newblock {Dialogue session segmentation by embedding-enhanced texttiling}.
\newblock \emph{arXiv preprint arXiv: 1610.03955}.

\bibitem[{Susnjak(2023)}]{susnjak2023applying}
Teo Susnjak. 2023.
\newblock {Applying BERT and ChatGPT for Sentiment Analysis of Lyme Disease in Scientific Literature}.
\newblock \emph{arXiv preprint arXiv: 2302.06474}.

\bibitem[{Wang et~al.(2021{\natexlab{a}})Wang, Song, Jiang, Lai, Yao, Zhang, and Su}]{ijcai2021-543}
Ante Wang, Linfeng Song, Hui Jiang, Shaopeng Lai, Junfeng Yao, Min Zhang, and Jinsong Su. 2021{\natexlab{a}}.
\newblock {A Structure Self-Aware Model for Discourse Parsing on Multi-Party Dialogues}.
\newblock In \emph{Proceedings of the Thirtieth International Joint Conference on Artificial Intelligence}, pages 3943--3949.

\bibitem[{Wang et~al.(2023{\natexlab{a}})Wang, Liang, Meng, Zou, Li, Qu, and Zhou}]{wang2023zeroshot}
Jiaan Wang, Yunlong Liang, Fandong Meng, Beiqi Zou, Zhixu Li, Jianfeng Qu, and Jie Zhou. 2023{\natexlab{a}}.
\newblock {Zero-Shot Cross-Lingual Summarization via Large Language Models}.
\newblock \emph{arXiv preprint arXiv: 2302.14229}.

\bibitem[{Wang et~al.(2023{\natexlab{b}})Wang, Lyu, Ji, Zhang, Yu, Shi, and Tu}]{wang2023documentlevel}
Longyue Wang, Chenyang Lyu, Tianbo Ji, Zhirui Zhang, Dian Yu, Shuming Shi, and Zhaopeng Tu. 2023{\natexlab{b}}.
\newblock {Document-Level Machine Translation with Large Language Models}.
\newblock \emph{arXiv preprint arXiv: 2304.02210}.

\bibitem[{Wang et~al.(2021{\natexlab{b}})Wang, Li, Zhao, and Yu}]{Wang_Li_Zhao_Yu_2021}
Xiaoyang Wang, Chen Li, Jianqiao Zhao, and Dong Yu. 2021{\natexlab{b}}.
\newblock {NaturalConv: A Chinese Dialogue Dataset Towards Multi-turn Topic-driven Conversation}.
\newblock In \emph{Proceedings of the AAAI Conference on Artificial Intelligence}, volume~35, pages 14006--14014.

\bibitem[{Wang et~al.(2023{\natexlab{c}})Wang, Xie, Ding, Feng, and Xia}]{wang2023chatgpt}
Zengzhi Wang, Qiming Xie, Zixiang Ding, Yi~Feng, and Rui Xia. 2023{\natexlab{c}}.
\newblock {Is ChatGPT a Good Sentiment Analyzer? A Preliminary Study}.
\newblock \emph{arXiv preprint arXiv: 2304.04339}.

\bibitem[{Wei et~al.(2022)Wei, Wang, Schuurmans, Bosma, brian ichter, Xia, Chi, Le, and Zhou}]{jason_wei_chain}
Jason Wei, Xuezhi Wang, Dale Schuurmans, Maarten Bosma, brian ichter, Fei Xia, Ed~Chi, Quoc~V Le, and Denny Zhou. 2022.
\newblock {Chain-of-Thought Prompting Elicits Reasoning in Large Language Models}.
\newblock In \emph{Proceedings of the 36th Conference on Neural Information Processing Systems}, pages 1--14.

\bibitem[{Wei et~al.(2023)Wei, Cui, Cheng, Wang, Zhang, Huang, Xie, Xu, Chen, Zhang, Jiang, and Han}]{wei2023zeroshot}
Xiang Wei, Xingyu Cui, Ning Cheng, Xiaobin Wang, Xin Zhang, Shen Huang, Pengjun Xie, Jinan Xu, Yufeng Chen, Meishan Zhang, Yong Jiang, and Wenjuan Han. 2023.
\newblock {Zero-Shot Information Extraction via Chatting with ChatGPT}.
\newblock \emph{arXiv preprint arXiv: 2302.10205}.

\bibitem[{Xie et~al.(2021)Xie, Liu, Xiong, Liu, and Copestake}]{xie-etal-2021-tiage-benchmark}
Huiyuan Xie, Zhenghao Liu, Chenyan Xiong, Zhiyuan Liu, and Ann Copestake. 2021.
\newblock {{TIAGE}: A Benchmark for Topic-Shift Aware Dialog Modeling}.
\newblock In \emph{Findings of the Association for Computational Linguistics: EMNLP 2021}, pages 1684--1690.

\bibitem[{Xing and Carenini(2021)}]{xing-carenini-2021-improving}
Linzi Xing and Giuseppe Carenini. 2021.
\newblock {Improving Unsupervised Dialogue Topic Segmentation with Utterance-Pair Coherence Scoring}.
\newblock In \emph{Proceedings of the 22nd Annual Meeting of the Special Interest Group on Discourse and Dialogue}, pages 167--177.

\bibitem[{Xing et~al.(2020)Xing, Hackinen, Carenini, and Trebbi}]{xing-etal-2020-improving}
Linzi Xing, Brad Hackinen, Giuseppe Carenini, and Francesco Trebbi. 2020.
\newblock {Improving Context Modeling in Neural Topic Segmentation}.
\newblock In \emph{Proceedings of the 1st Conference of the Asia-Pacific Chapter of the Association for Computational Linguistics and the 10th International Joint Conference on Natural Language Processing}, pages 626--636.

\bibitem[{Xu et~al.(2021)Xu, Zhao, and Zhang}]{Xu_Zhao_Zhang_2021}
Yi~Xu, Hai Zhao, and Zhuosheng Zhang. 2021.
\newblock {Topic-aware Multi-turn Dialogue Modeling}.
\newblock In \emph{Proceedings of the AAAI Conference on Artificial Intelligence}, volume~35, pages 14176--14184.

\bibitem[{Yang et~al.(2022)Yang, Lin, Li, Meng, Wang, Wang, and Zhou}]{yang-etal-2022-take}
Chenxu Yang, Zheng Lin, Jiangnan Li, Fandong Meng, Weiping Wang, Lanrui Wang, and Jie Zhou. 2022.
\newblock {{TAKE}: Topic-shift Aware Knowledge s{E}lection for Dialogue Generation}.
\newblock In \emph{Proceedings of the 29th International Conference on Computational Linguistics}, pages 253--265.

\bibitem[{Yang et~al.(2021)Yang, Xu, Xu, Li, Gao, Guo, Xue, and Wen}]{yang2021joint}
Jingxuan Yang, Kerui Xu, Jun Xu, Si~Li, Sheng Gao, Jun Guo, Nianwen Xue, and Ji-Rong Wen. 2021.
\newblock {A Joint Model for Dropped Pronoun Recovery and Conversational Discourse Parsiin Chinese Conversational Speech}.
\newblock In \emph{Proceedings of the 59th Annual Meeting of the Association for Computational Linguistics and the 11th International Joint Conference on Natural Language Processing}, pages 1752--1763.

\bibitem[{Yang et~al.(2023)Yang, Li, Zhang, Chen, and Cheng}]{yang2023exploring}
Xianjun Yang, Yan Li, Xinlu Zhang, Haifeng Chen, and Wei Cheng. 2023.
\newblock {Exploring the Limits of ChatGPT for Query or Aspect-based Text Summarization}.
\newblock \emph{arXiv preprint arXiv: 2302.08081}.

\bibitem[{Yu et~al.(2022)Yu, Fu, and Zhang}]{yu2022speaker}
Nan Yu, Guohong Fu, and Min Zhang. 2022.
\newblock {Speaker-Aware Discourse Parsing on Multi-Party Dialogues}.
\newblock In \emph{Proceedings of the 29th International Conference on Computational Linguistics}, pages 5372--5382.

\bibitem[{Yuan et~al.(2023)Yuan, Xie, and Ananiadou}]{yuan2023zeroshot}
Chenhan Yuan, Qianqian Xie, and Sophia Ananiadou. 2023.
\newblock {Zero-shot Temporal Relation Extraction with ChatGPT}.
\newblock \emph{arXiv preprint arXiv: 2304.05454}.

\bibitem[{Zhang et~al.(2023{\natexlab{a}})Zhang, Liu, and Zhang}]{zhang2023extractive}
Haopeng Zhang, Xiao Liu, and Jiawei Zhang. 2023{\natexlab{a}}.
\newblock {Extractive Summarization via ChatGPT for Faithful Summary Generation}.
\newblock \emph{arXiv preprint arXiv: 2304.04193}.

\bibitem[{Zhang et~al.(2018)Zhang, Dinan, Urbanek, Szlam, Kiela, and Weston}]{zhang-etal-2018-personalizing}
Saizheng Zhang, Emily Dinan, Jack Urbanek, Arthur Szlam, Douwe Kiela, and Jason Weston. 2018.
\newblock {Personalizing Dialogue Agents: {I} have a dog, do you have pets too?}
\newblock In \emph{Proceedings of the 56th Annual Meeting of the Association for Computational Linguistics}, pages 2204--2213.

\bibitem[{Zhang et~al.(2023{\natexlab{b}})Zhang, Ladhak, Durmus, Liang, McKeown, and Hashimoto}]{zhang2023benchmarking}
Tianyi Zhang, Faisal Ladhak, Esin Durmus, Percy Liang, Kathleen McKeown, and Tatsunori~B. Hashimoto. 2023{\natexlab{b}}.
\newblock {Benchmarking Large Language Models for News Summarization}.
\newblock \emph{arXiv preprint arXiv: 2301.13848}.

\end{thebibliography}

% \section*{Language Resource References}
% \label{lr:ref}
% \bibliographystylelanguageresource{lrec-coling2024-natbib}
% \bibliographylanguageresource{languageresource}

\appendix

\section{Appendix}
\label{sec:reference}
\subsection{Details of Human Evaluation for Topic Structures}
\label{detailsofhumanevaluation}
To compare the topic structures annotated by humans and ChatGPT, we recruited 3 annotators whose native language is Chinese. All of the annotators are undergraduate students studying at a university where English is the official language. Each annotator is instructed to compare the topic structures annotated by ChatGPT and humans in order to determine which one is more reasonable. The model names remain anonymous, and the positions of the model outputs are randomly swapped. We finally adopted voting to avoid individual bias.   
\subsection{Details of the Variants of Various Components in Prompt}
\label{detailsofvariants}
The components of the prompt we designed mainly consist of task description, output format, and structured input. We introduce the variants of each component in detail below.
\begin{table*}[]
\centering
\begin{tabular}{l}
\hline
\multicolumn{1}{c}{\textbf{Prompt}}                                                                                                                                                                                                                                                                                                                                                                    \\ \hline
\begin{tabular}[l]{@{}l@{}}  \begin{tabular}[l]{@{}l@{}} The following is a description to guide the generative model to complete the dialogue topic segmentation \\ task/dialogue discourse parsing task. \\ \textless{}Task description written by humans\textgreater\\  Please generate a similar description with several sentences.\end{tabular}    \\ \hline   \end{tabular}
\end{tabular}
\caption{Prompt that the task description be generated using ChatGPT. }
\label{prompt2chatgptGenerated}
\end{table*}
\paragraph{Task Description}
The task description mainly includes two types: human-written and ChatGPT - generated, which guides ChatGPT to complete the task as required.
The descriptions written by humans are shown in Table~\ref{promptsDDA}.  For each task, we describe the goal of the task, such as identifying several boundaries for dialogue topic segmentation, to instruct ChatGPT to understand and complete the task.
In addition, to get the task description generated by ChatGPT, we feed the task description written by humans to ChatGPT and instruct ChatGPT generate a similar description, and the prompt is shown in Table~\ref{prompt2chatgptGenerated}.

\paragraph{Output Format}
Output format for dialogue topic segmentation mainly includes two forms: python dictionary and sequence labeling. 
For example, given a dialogue consisting of 10 utterances, denoted as $\{u_1,u_2,u_3,\cdots, u_{10}\}$, where $u_1-u_4$, $u_5-u_8$ and $u_9-u_{10}$ are considered different topics. Python dictionary form is as {'topic 1': [1, 2, 3, 4], 'topic 2': [5, 6, 7, 8], 'topic 3': [9, 10]}, where the elements in the list are the index of the consecutive utterances within the topic. Sequence labeling form is a list of length 10 containing 0 or 1, where 1 indicates the end of the topic, as follows $[0, 0, 1, 0, 0, 1, 0, 1]$.

Output format for dialogue discourse parsing mainly includes sparse matrix and adjacency matrix forms.
For example, given a dialogue consisting of 3 utterances, denoted as $\{u_1, u_2, u_3\}$, where $u_2$ and $u_1$ form the Question-answer pair type, and $u_3$ and $u_1$ form the Clarification\_question type. The sparse matrix form is as  [[1, 2 Question-answer pair], [1, 3, Clarification\_question]], where the elements in the list are the indexes of two utterances, and the relation type. The adjacency matrix form is an adjacency matrix of shape $3\times3$, as follows [[0, 0, 0], [Question-answer\_pair, 0, 0], [Clarification\_question, 0, 0]], where the element $w_{ij}$ in the adjacency matrix is the relation type between $u_i$ and $u_j$.

\paragraph{Input Format}
Input format include structured and unstructured types, referring to whether or not feed the utterances utterance to ChatGPT line by line.
For example, given a dialogue consisting of 3 utterances, denoted as $\{u_1, u_2, u_3\}$. The structured form is as "1: $u_1$ $\backslash$n 2: $u_2$ $\backslash$n 3: $u_3$: $\backslash$n", while unstructured form means removing these utterance indicators.
\end{document}